\documentclass[journal,twoside,web]{ieeecolor}
\usepackage{jsen}
\usepackage{cite}
\usepackage{amsmath,amssymb,amsfonts}
\usepackage{algorithm,algorithmic}
\usepackage{multicol,multirow}
\usepackage{graphicx}
\usepackage{subcaption}
\usepackage{textcomp}
\usepackage{wrapfig}
\usepackage{hyperref}
\hypersetup{
    colorlinks=true,
    linkcolor=black,
    citecolor=black,
    urlcolor=black
}

\def\BibTeX{{\rm B\kern-.05em{\sc i\kern-.025em b}\kern-.08em
    T\kern-.1667em\lower.7ex\hbox{E}\kern-.125emX}}
\definecolor{abstractbg}{rgb}{0.89804,0.94510,0.83137}
\begin{document}

\title{Spatiotemporal EEG-Based Emotion Recognition Using SAM Ratings from Serious Games with Hybrid Deep Learning}

\author{Abdul Rehman, Ilona Heldal, and Jerry Chun-Wei Lin
\thanks{A. Rehman, I. Heldal, and J. C.-W. Lin are with the Department of Computer Science, Electrical Engineering and Mathematical Sciences, Western Norway University of Applied Sciences, 5020 Bergen, Norway (e-mail: \{arj, ilona.heldal, jerry.chun-wei.lin\}@hvl.no).}
\thanks{Corresponding author: Abdul Rehman, Jerry Chun-wei Lin (e-mail: arj@hvl.no, jerry.chun-wei.lin\}@hvl.no).}
}

\IEEEtitleabstractindextext{%
\fcolorbox{abstractbg}{abstractbg}{%
\begin{minipage}{\textwidth}%
\begin{wrapfigure}[5.5]{r}{3in}%
\vspace{-10pt}
\includegraphics[width=3in]{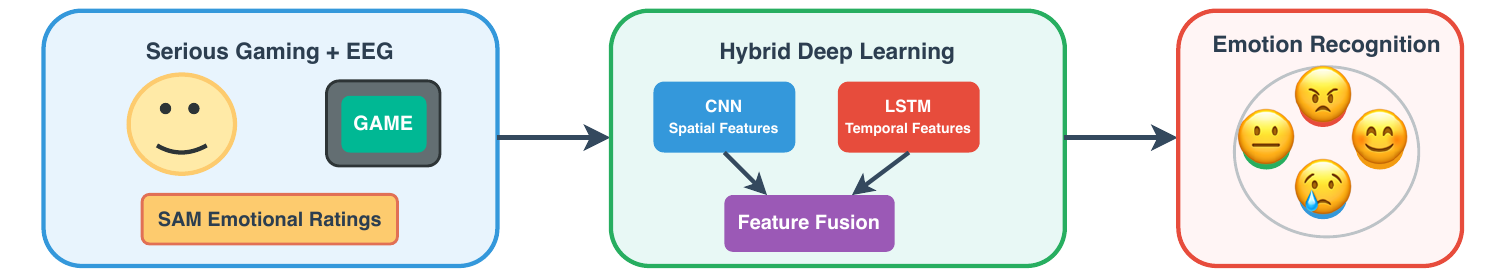}%
\end{wrapfigure}%
\begin{abstract}
Recent advancements in EEG-based emotion recognition have shown promising outcomes using both deep learning and classical machine learning approaches; however, most existing studies focus narrowly on binary valence prediction or subject-specific classification, which limits generalizability and deployment in real-world affective computing systems. To address this gap, this paper presents a unified, multigranularity EEG emotion classification framework built on the GAMEEMO dataset, which consists of 14-channel EEG recordings and continuous self-reported emotion ratings (boring, horrible, calm, and funny) from 28 subjects across four emotion-inducing gameplay scenarios. Our pipeline employs a structured preprocessing strategy that comprises temporal window segmentation, hybrid statistical and frequency-domain feature extraction, and z-score normalization to convert raw EEG signals into robust, discriminative input vectors. Emotion labels are derived and encoded across three complementary axes: (i) binary valence classification based on the averaged polarity of positive (funny, calm) and negative (boring, horrible) emotion ratings, and (ii) Multi-class emotion classification, where the presence of the most affective state is predicted. (iii) Fine-grained multi-label representation via binning each emotion into 10 ordinal classes. We evaluate a broad spectrum of models, including Random Forest, XGBoost, and SVM, alongside deep neural architectures such as LSTM, LSTM-GRU, and CNN-LSTM. Among these, the LSTM-GRU model consistently outperforms the others, achieving an F1-score of 0.932 in the binary valence task and 94.5\% and 90.6\% in both multi-class and Multi-Label emotion classification. In contrast to prior work, our framework delivers high-resolution affect modeling with generalized classification capacity and strong subject-independent reproducibility, offering a scalable solution for future real-time EEG-based emotion recognition applications.
\end{abstract}

\end{minipage}}}

\maketitle

\section{Introduction}
\label{sec:introduction}
\IEEEPARstart{E}{motion} recognition plays a critical role in affective computing that enables various intelligent systems to recognize, infer, and interpret interpersonal interactions, knowledge insight, perception, and human reactions/states to improve user experiences as well as enhance their outcomes \cite{wang2022transformers,shahzad2024eeg,gosala2024hybrid}. Recent advances in Artificial Intelligence (AI) have significantly enhanced the ability to detect and interpret emotions, revolutionizing fields such as human-computer interaction \cite{li2022eeg,saxena2020emotion} and mental health \cite{olawade2024enhancing}. Notably, deep learning models have enhanced the accuracy of emotion detection by effectively analyzing complex data, such as facial expressions, voice tones, and text \cite{vistorte2024integrating,bekenova2023emotion}. Transfer learning has further accelerated progress by enabling models trained on large datasets to be adapted to specific tasks with smaller datasets, improving generalization across different populations and contexts \cite{bragilovski2023tltd}.

Emotion recognition plays a critical role in affective computing, enabling the interpretation of human reactions \cite{chen2021emotion}. Emotions vary significantly across different age groups due to psychological, social, and physiological changes that occur throughout a person's life. As children learn to navigate new experiences and manage their emotions, they often exhibit rapid fluctuations between different emotional states, such as happiness, anger, or melancholy \cite{albraikan2018toward,sesso2024reactivity}. Adults are better at controlling their emotions and understanding them more deeply, even though they still occasionally feel stressed or anxious about their life obligations. The Elderly often experience changes in emotional regulation and expression as they age, which can be influenced by life experiences, cognitive changes, and social circumstances \cite{yadav2018emotional}. For instance, older adults may express emotions more subtly or with different facial and vocal cues compared to younger adults. The emotional priorities of older people may also change; they may exhibit less severe outward manifestations of unpleasant emotions, such as sadness or anger, instead choosing to concentrate more on preserving social ties and positive emotions. 

While there exist some studies on emotion recognition, they are often limited to only identifying and categorizing emotion \cite{kamble2021ensemble,shahzad2024eeg,gosala2024hybrid}. Since different emotions are difficult to generalize, they cannot work effectively on datasets that are too diverse or too large \cite{dunsmoor2015fear}. Most existing studies have focused on multiclass classification (4 classes) only. In this paper, we build upon existing research, considering both binary classification and multi-label classification. This paper makes the following contributions:

\begin{itemize}
\item We present a unified, multigranularity EEG emotion classification framework built on the GAMEEMO dataset, which consists of 14-channel EEG recordings and continuous self-reported emotion ratings (boring, horrible, calm, and funny) from 28 subjects across four emotion-inducing gameplay scenarios. 
\item Our pipeline employs a structured preprocessing strategy that comprises temporal window segmentation, hybrid statistical and frequency-domain feature extraction, and z-score normalization to convert raw EEG signals into robust, discriminative input vectors. Emotion labels are derived and encoded across three complementary axes: (i) binary valence classification based on the averaged polarity of positive (funny, calm) and negative (boring, horrible) emotion ratings, and (ii) Multi-class emotion classification, where the presence of the most affective state is predicted. (iii) Fine-grained multi-label representation via binning each emotion into 10 ordinal classes.
\item  We evaluate a broad spectrum of models, including Random Forest, XGBoost, and SVM, alongside deep neural architectures such as LSTM, LSTM-GRU, and CNN-LSTM. Among these, the LSTM-GRU model consistently outperforms the others, achieving an F1-score of 0.932 in the binary valence task and 94.5\% and 90.6\% in both multi-class and Multi-Label emotion classification. In contrast to prior work, our framework delivers high-resolution affect modeling with generalized classification capacity and strong subject-independent reproducibility, offering a scalable solution for future real-time EEG-based emotion recognition applications.
\end{itemize}

This paper is organized as follows: Section \ref{RW} provides the related work on emotion recognition using AI. Utilized ML and DL models, data preprocessing methods, dataset preliminaries, and the technical components of the proposed framework are all covered in detail in Section \ref{prop}. In Section \ref{EAR}, the experimental analysis provides a thorough examination and draws conclusions based on the proposed work. Section \ref{Discussion} then presents the discussion. Finally, Section \ref{con} concludes the paper and leads to future directions.

\section{Related Work}\label{RW}
This section provides the related work on emotion recognition. EEG emotion detection research has been exciting for a long time, and several papers have proposed different approaches to creating systems that can identify emotions in humans. 

Shahzad et al. \cite{shahzad2024eeg} focused on using EEG data from 28 subjects, utilizing the GAMEEMO dataset, for emotion recognition in a gaming environment. They developed an EEG-driven framework for emotion recognition during gameplay, utilizing Random Forest to detect emotions (boring, calm, horror, and funny) with an accuracy of 98.21\%. This framework is validated on EEG data collected during gaming, demonstrating its efficacy in recognizing emotional states with applications in adaptive gaming and affective computing. Gosala et al. \cite{gosala2024hybrid} introduced a hybrid convolutional neural network (CNN) model utilizing the GAMEEMO extended dataset, which comprises EEG signals recorded during gameplay of four emotionally distinct games (boring, calm, horror, and funny). They highlight the effectiveness of hybrid CNN architectures in processing EEG data for emotion detection. Chen et al. \cite{chen2021emotion} presented a multimodal emotion identification approach using a combination of EEG and ECG and the Dempster-Shafer evidence theory. The SVM classifier is utilised for EEG feature classification. They utilise the bi-directional long short-term memory for ECG. Superior performance over single-modal models is achieved by fusing the outcomes using evidence theory. By 2.64\% in arousal and 2.75\% in valence, the multimodal model outperforms EEG-based models; in comparison, it outperforms ECG-based models by 7.37\% and 8.73\%, respectively. An enhanced method for EEG-based emotion recognition on the publicly accessible VREED dataset was presented in \cite{uyanik2022use}. Two emotional states were automatically classified by five machine learning classifiers using DE characteristics.
Additionally, they observed that the gamma band yielded the highest average accuracy score. Alslaity et al. \cite{alslaity2024machine} examined emotion recognition in a virtual reality setting using machine learning and explainable machine learning methods. Fang et al. \cite{fang2021multi} recognized emotions using EEG data by employing a multi-feature input deep forest (MF-DF) model, which analyzed the DEAP dataset's two valence and arousal labels. Ahirwal et al. \cite{ahirwal2020audio} introduced a novel channel selection method for emotion categorisation using EEG signals produced by audio-visual stimulation from 40 participants. Three methods are used: artificial neural networks (ANN), naive bayes (NB), and SVM. Diah et al. \cite{diah2019exploring} created an EEG-based emotion identification system with a feature extraction and classifier subsystem. They employ and analyze nine characteristics extracted from the EEG signal's temporal and frequency domains. To categorise the subject's emotional state, they applied Random Forest and SVM techniques, and they contrasted the outcomes with those of other machine learning techniques. The results of the experiment indicate that the Random Forest approach yields the maximum recognition accuracy, measuring 62.58\%.

Algumaei et al. \cite{algumaei2021wavelet} introduced a model for emotion recognition using wavelet packet energy characteristics and EEG data. The emotional states were detected using four conventional classifiers: naive Bayes, k-nearest neighbor, SVM, and linear discriminant analysis. When listening to bilingual (English and Urdu) music, Zainba et al. \cite{zainab2021emotion} used EEG signals to identify four distinct emotions: joyful, sad, angry, and calm. To create the hybrid feature vector, which classifiers then use to identify emotional reactions, frequency and time-domain characteristics are merged. Happiness is the most prevalent and easily identifiable emotion, and hybrid characteristics have been found to produce superior outcomes than separate domains. Bardak et al. \cite{bardak2024adaptive} aimed to improve the recognition accuracy of three and four emotions by using the Neuro-Fuzzy Inference System (ANFIS). The investigation's results demonstrate that the proposed model efficiently detects emotions and that it performs well in terms of categorization. Zhong et al. \cite{zhong2020eeg} proposed a Regularised Graph Neural Network (RGNN) using an EEG to recognize emotions. It suggests two regularizers to better handle noisy labeling and cross-subject EEG variability. Three emotion classes are used to thoroughly analyze the proposed RGNN model: neutral, negative, and positive.

\section{Proposed Methodology} \label{prop}
\begin{figure*}[!ht]
 \centering
 \includegraphics[width=\textwidth]{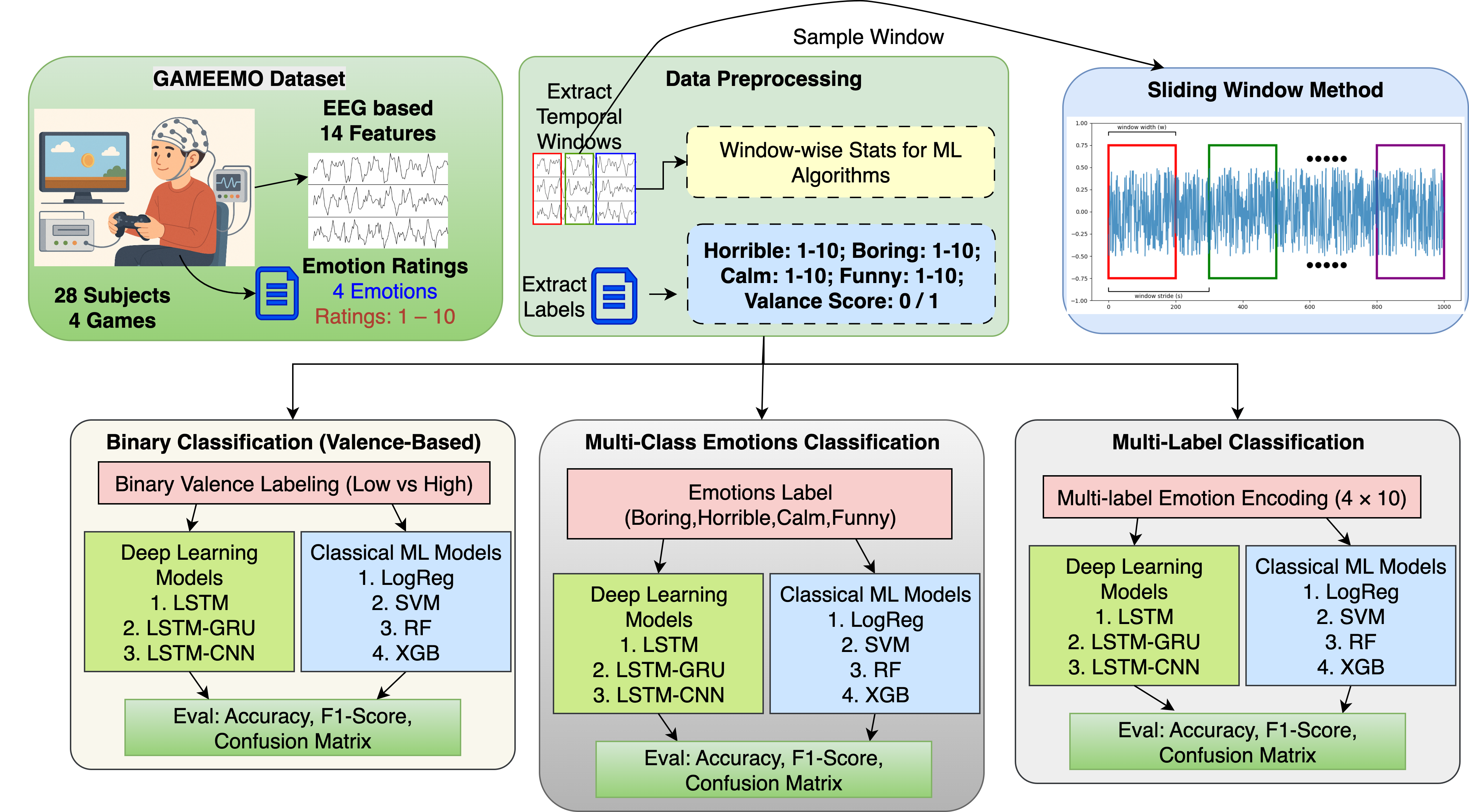}
\caption{Proposed framework for EEG-based emotion classification leveraging the GAMEEMO dataset.}
\label{Fig.1}
\end{figure*}

\begin{figure*}[!ht]
 \centering
 \includegraphics[width=\textwidth]{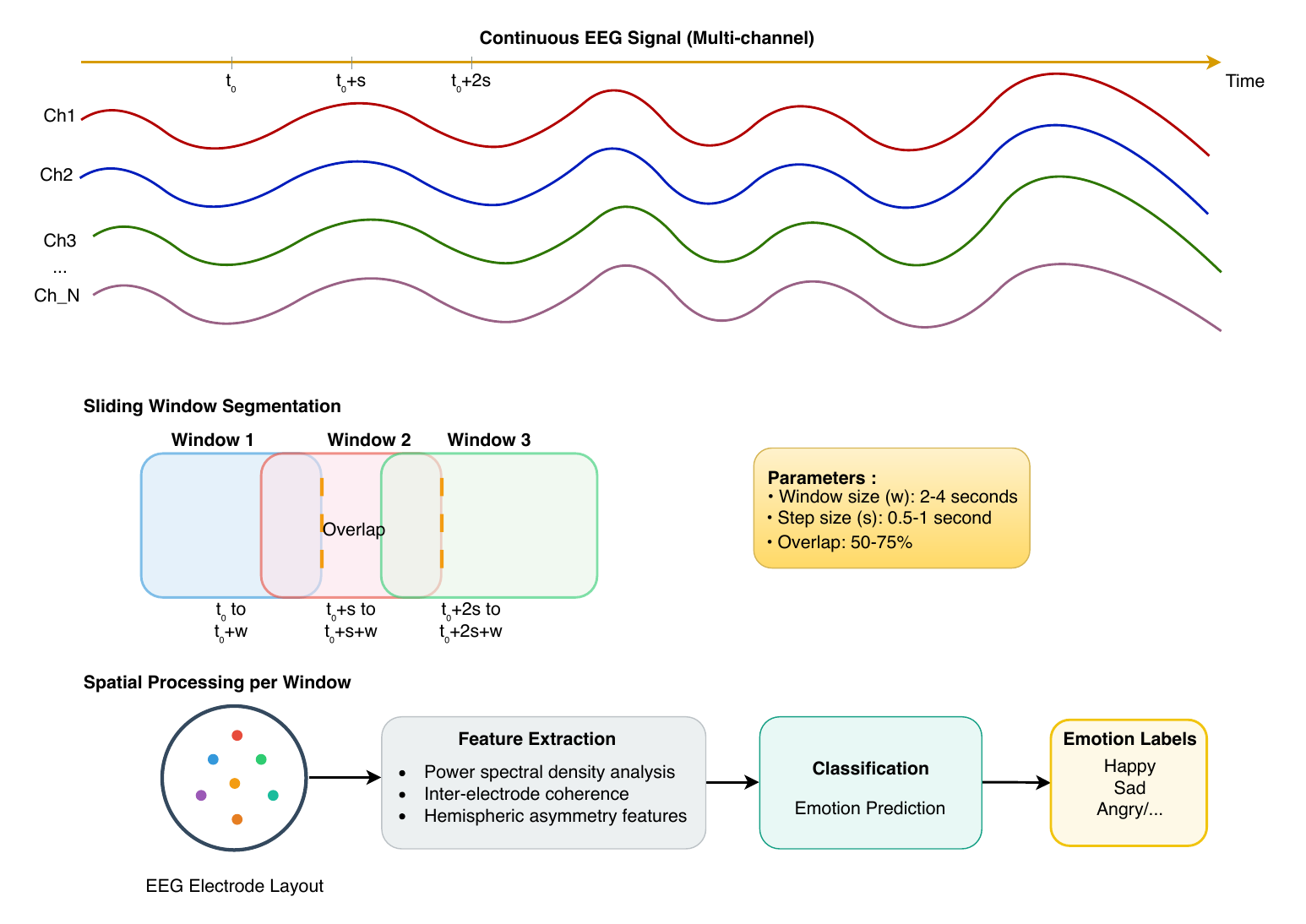}
\caption{High Level Illustration of sliding window for Spatiotemporal EEG-based Emotion Recognition}
\label{Fig.High}
\end{figure*}

Figure~\ref {Fig.1} presents the overall architecture of the proposed EEG-based emotion classification framework designed for cognitive assessment during gameplay, leveraging the structured GAMEEMO dataset.
This dataset comprises EEG signals from 28 participants across four gameplay sessions, with each EEG record containing 14 structured features per time point and accompanied by self-reported emotion ratings (scale 1–10) for four affective states: horrible, boring, calm, and funny. To address the non-stationary nature of EEG data, a comprehensive preprocessing pipeline is employed. First, a sliding window technique segments the continuous EEG time series into overlapping epochs, enabling both temporal feature preservation and efficient batch processing. For each window, a range of statistical descriptors, such as mean, standard deviation, and entropy, is extracted to construct robust input vectors. Fig. \ref{Fig.High} Corresponding emotion labels are transformed into three task-specific formats: (i) binary valence classification based on the averaged polarity of positive (funny, calm) and negative (boring, horrible) emotion ratings, and (ii) Multi-class emotion classification, where the presence of the most affective state is predicted. (iii) Fine-grained multi-label representation via binning each emotion into 10 ordinal classes. These representations feed into parallel classification pipelines comprising both deep learning architectures (LSTM, LSTM-GRU, LSTM-CNN) and classical machine learning models (Logistic Regression, SVM, Random Forest, XGBoost). Each path is independently trained and evaluated using metrics such as accuracy, F1-score, and confusion matrix analysis. This modular and extensible framework supports the detailed decoding of emotions from EEG, enabling both fine-grained and generalized predictions of emotional states across diverse affective dimensions.

Algorithm~\ref{algo1} outlines a unified and structured five-phase framework designed for EEG-based emotion recognition using the GAMEEMO dataset, targeting both multi-label multi-class and binary valence classification objectives. The pipeline begins by segmenting raw EEG signals collected from 28 subjects across four gameplay sessions into overlapping temporal windows (500 ms, 50\% overlap), thereby preserving temporal dynamics while standardizing input lengths for downstream processing. From each window, a rich set of statistical (mean, variance, entropy) and frequency-domain features (e.g., FFT, bandpower) is extracted to form discriminative representations of brain activity. These feature vectors are then normalized using z-score standardization to minimize inter-subject and inter-session variability. Emotion annotations, derived from synchronized PDF-based self-reports, are transformed into two distinct supervised targets: a multi-class format with 10-level binning for each of the four emotions (boring, horrible, calm, funny), and a binary valence label, computed by thresholding the average intensity of positive (funny, calm) versus negative (boring, horrible) affective scores. These labeled datasets feed into parallel learning pipelines consisting of both deep models (LSTM, LSTM-GRU, LSTM-CNN) and classical machine learning algorithms (Logistic Regression, SVM, Random Forest, XGBoost). Each branch is optimized using the AdamW optimizer with cosine annealing learning rate scheduling and warm-up strategy, along with regularization techniques such as dropout (0.3), weight decay ($10^{-4}$), and gradient clipping (norm $\leq$ 1.0). Model training spans 100 epochs with a batch size of 32, logging accuracy, precision, recall, F1-score, and confusion matrices per epoch for comprehensive evaluation. Final models are evaluated on held-out test sets, and outputs, along with visualizations, are exported via TensorBoard or saved as high-resolution plots. This modular and scalable design facilitates robust emotion decoding from EEG signals, supporting both fine-grained modeling of emotional states and high-level valence analysis.

\begin{algorithm}[!ht]
\caption{EEG-Based Emotion Recognition via Multi-Label, Binary, and Multi-Label Emotion Classification}
\label{algo1}
\begin{algorithmic}[1]
\REQUIRE EEG signals $\mathbf{X} \in \mathbb{R}^{N \times C \times T}$, emotion labels $\mathbf{Y} \in \mathbb{R}^{N \times 4}$ from 28 subjects across 4 games

\STATE \textbf{Phase 1: Signal Preprocessing}
\STATE Segment $\mathbf{X}$ into overlapping windows $\mathbf{W}_i \in \mathbb{R}^{C \times \tau}$, with $\tau = 500\,ms$ and $50\%$ overlap
\STATE Extract features: $\mathbf{F}_i = f_{\text{stat+freq}}(\mathbf{W}_i)$
\STATE Normalize: $\hat{\mathbf{F}}_i = \text{zscore}(\mathbf{F}_i)$

\STATE \textbf{Phase 2: Label Construction}
\STATE \textit{Multi-class:} $\mathbf{Y}_{ml} = \text{bin}_{10}(\mathbf{Y}) \in \{0,\dots,10\}^4$
\STATE \textit{Binary valence:} $y_{bin} = \mathbb{1}\left[\text{mean}(y_{\text{funny}}, y_{\text{calm}}) > \text{mean}(y_{\text{boring}}, y_{\text{horrible}})\right]$
\STATE \textit{Multi-Label Emotion (categorical):} $y_{\text{cat}} \in \{\text{boring}, \text{horrible}, \text{calm}, \text{funny}\}$ via argmax

\STATE \textbf{Phase 3: Model Definitions}
\STATE Define model sets $\mathcal{M}_{ml}$, $\mathcal{M}_{bin}$, $\mathcal{M}_{cat}$ for each task
\STATE \quad Deep models: LSTM, GRU, LSTM-CNN \quad Classical: LR, SVM, RF, XGB
\STATE \quad Loss: BCE (multi-class), CE (binary, categorical)
\STATE \quad Optimizer: AdamW ($\eta=0.005$) + cosine scheduler + warm-up
\STATE \quad Regularization: dropout $=0.3$, weight decay $=10^{-4}$, gradient clip $\leq 1.0$

\STATE \textbf{Phase 4: Training and Validation}
\FOR{$e = 1$ to $E = 100$}
    \FOR{$\mathcal{M} \in \{\mathcal{M}_{ml}, \mathcal{M}_{bin}, \mathcal{M}_{cat}\}$}
        \STATE Train $\mathcal{M}$ on $\hat{\mathbf{F}}_{\text{train}}$ (batch size = 32)
        \STATE Validate on $\hat{\mathbf{F}}_{\text{val}}$, compute $\text{Acc}, F_1, \text{Prec}, \text{Rec}$
        \STATE Save $\text{ConfMatrix}_{\mathcal{M}}$
    \ENDFOR
\ENDFOR

\STATE \textbf{Phase 5: Testing and Export}
\STATE Evaluate $\mathcal{M}$ on held-out test set, compute per-class metrics
\RETURN Trained models $\{\mathcal{M}_{ml}, \mathcal{M}_{bin}, \mathcal{M}_{cat}\}$
\end{algorithmic}
\end{algorithm}

\subsection{Dataset Collection and Pre-processing}
The proposed EEG-based emotion classification framework utilizes the structured GAMEEMO dataset, which comprises EEG recordings from 28 participants participating in four emotionally evocative gaming scenarios. Each session is annotated with self-reported emotion scores for four affective states: boring, horrible, calm, and funny, rated on a continuous scale from 0 to 10. EEG data, recorded across 14 channels, undergoes a robust multi-stage pre-processing pipeline to retain temporal dynamics and enhance feature discriminability. Initially, raw signals are segmented into overlapping temporal windows (e.g., 500 ms with 50\% overlap), allowing for localized temporal analysis of neural fluctuations. Within each segment, comprehensive handcrafted features are extracted per channel, including time-domain statistics (mean, standard deviation, entropy, skewness, kurtosis) and frequency-domain features (band powers across delta, theta, alpha, and beta bands), forming a high-dimensional representation of the subject’s cognitive-emotional state at fine granularity. This feature extraction process follows structured principles similar to those in physics-constrained data-driven modeling, where domain-specific transformations improve representation fidelity and learning robustness ~\cite{bao2019physics}.
Labels are constructed using two parallel strategies: (1) in the multi-class setting, each emotion is discretized into 11 ordinal bins (0–10), resulting in a multi-hot 4-dimensional label vector that supports nuanced modeling of co-occurring affective states; and (2) in the binary classification setup, a valence score is computed by averaging positive (funny, calm) and negative (boring, horrible) ratings, followed by thresholding to assign a binary label reflecting emotional polarity. To ensure consistency and model convergence, z-score normalization is applied to all features, and padding is used where necessary to maintain uniform segment lengths. Additionally, the framework accommodates a Multi-Label Emotion classification mode, where all four discrete emotions are predicted simultaneously, capturing the interdependencies and overlapping patterns across affective states, thus enabling a richer representation of complex emotional responses. The dataset is then partitioned using an 80:20 subject-independent split to maintain generalization integrity, ensuring subjects in the validation set remain unseen during training and that class distributions are preserved for all classification paradigms.
\section{Experimental Analysis, Results, and Discussion}\label{EAR}
The comparative results in Table.~\ref{Table.1} highlight the superior performance of deep learning models, particularly the LSTM-GRU architecture, across all three classification paradigms: binary valence, multi-class emotion recognition, and Multi-Label Emotion classification. In the binary setup where emotional valence is computed based on the balance between positive and negative emotions, the LSTM-GRU model achieved a leading F1-score of 0.932 and an accuracy of 93.3\%, notably outperforming the classical Random Forest model, which attained an accuracy of 85\%. For the multi-class task, which independently predicts the four emotions "boring," "horrible," "calm," and "funny," LSTM-GRU again led with F1-scores above 0.92 for each class. At the same time, the best-performing Random Forest model yielded F1-scores between 0.738 and 0.792, suggesting challenges in learning finer affective nuances. In the more complex Multi-Label Emotion classification scenario, where the model must simultaneously determine the dominant emotional state from a set of four classes, the LSTM-GRU model achieved 90.6\% accuracy and an F1-score of 0.909, substantially outperforming the Random Forest's 79\%. These results collectively validate the importance of temporal sequence modeling and feature-rich preprocessing in the fine-grained decoding of affective EEG signals. The deep models consistently demonstrated stronger generalization and robustness across subject-independent evaluations, making them ideal candidates for real-time EEG-based emotion recognition systems.

\begin{table*}[!ht]
\centering
\caption{Performance Comparison of Top-Performing ML and DL Models on Binary, Multi-Class, and Multi-Label Emotion EEG-Based Classification Tasks}
\label{Table.1}
\begin{tabular}{|c|c|c|c|c|c|}
\hline
\textbf{Task} & \textbf{Model Type} & \textbf{Emotion Class} & \textbf{Accuracy} & \textbf{Precision} & \textbf{F1-Score} \\
\hline
\multirow{2}{*}{Binary (Valence)} 
 & ML - Random Forest & Positive / Negative & 0.85 & 0.85 (avg) & 0.85 (avg) \\
 & DL - LSTM-GRU & Positive / Negative & \textbf{0.933} & \textbf{0.933} & \textbf{0.932} \\
\hline
\multirow{10}{*}{Multi-Class (Per Emotion)} 
 & \multirow{5}{*}{ML - Random Forest} 
 & boring & 0.808 & 0.892 & 0.740 \\
 & & horrible & 0.796 & 0.841 & 0.738 \\
 & & calm & 0.791 & 0.832 & 0.763 \\
 & & funny & 0.790 & 0.808 & 0.792 \\
 & & \textbf{Macro Average} & \textbf{0.796} & \textbf{0.843} & \textbf{0.758} \\
 \cline{2-6}
 & \multirow{5}{*}{DL - LSTM-GRU} 
 & boring & \textbf{0.955} & \textbf{0.942} & \textbf{0.939} \\
 & & horrible & \textbf{0.953} & \textbf{0.948} & \textbf{0.943} \\
 & & calm & \textbf{0.938} & \textbf{0.929} & \textbf{0.927} \\
 & & funny & \textbf{0.934} & \textbf{0.934} & \textbf{0.935} \\
 & & \textbf{Macro Average} & \textbf{0.945} & \textbf{0.938} & \textbf{0.936} \\
\hline

\multirow{2}{*}{Multi-Label (Joint Emotion Labels)} 
 & ML - Random Forest & boring, horrible, calm, funny (1-10) & 0.79 & 0.79 & 0.79 \\
 & DL - LSTM-GRU & boring, horrible, calm, funny (1-10) & \textbf{0.906} & \textbf{0.909} & \textbf{0.909} \\
\hline
\end{tabular}
\end{table*}

\begin{figure}[!ht]
\centering
\subfloat[Random Forest\label{binary_cma}]{
\includegraphics[width=0.49\columnwidth]{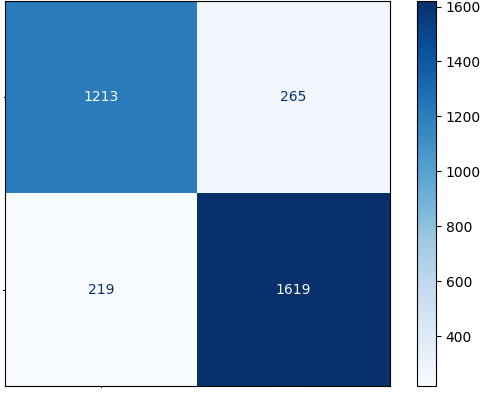}}
\subfloat[LSTM-GRU\label{binary_cmb}]{
\includegraphics[width=0.49\columnwidth]{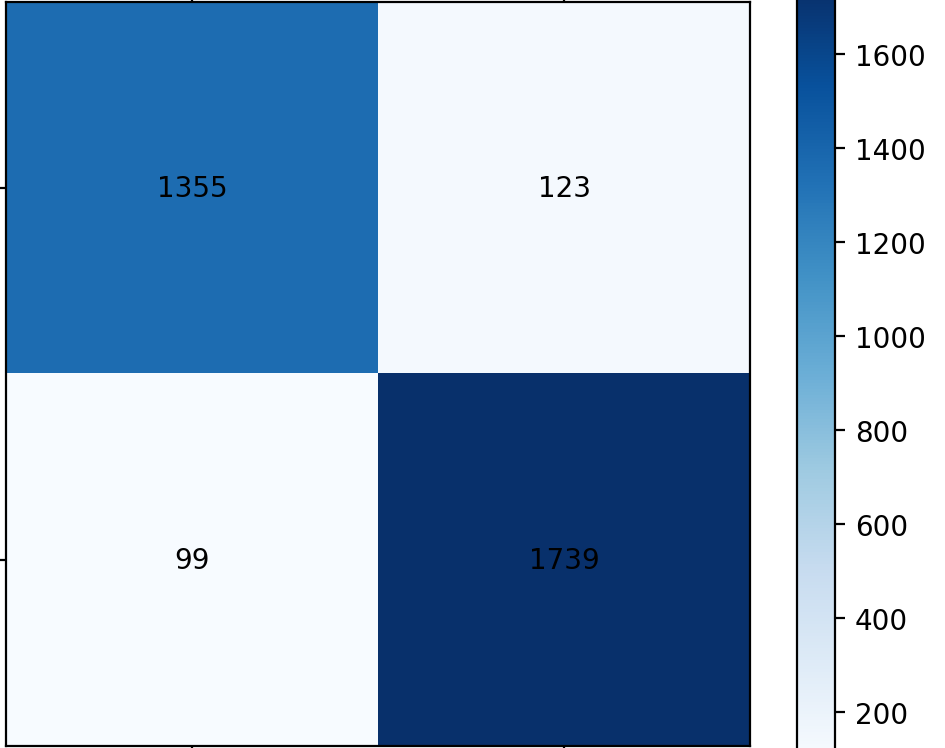}}
\caption{Confusion matrices for Binary Classification}
\label{fig:binary_cm}
\end{figure}

This paragraph presents a comprehensive visualization of confusion matrices for the highest-performing models across all three classification tasks: binary valence detection, multi-class emotion recognition, and multi-label emotion classification. In the binary classification task, the Random Forest model (Fig.~\ref{binary_cma}) produces balanced and moderately accurate predictions, characterized by a relatively dense diagonal and few off-diagonal misclassifications, indicating a solid separation between low and high valence states. The LSTM-GRU model (Fig.~\ref{binary_cmb}), however, further strengthens this separation by yielding a sharper diagonal and fewer false positives and negatives, highlighting its ability to capture fine-grained temporal dependencies in EEG data.

\begin{figure}[!ht]
\centering
\subfloat[Random Forest \label{multi_class_emotion_cma}]{
\includegraphics[width=0.49\columnwidth]{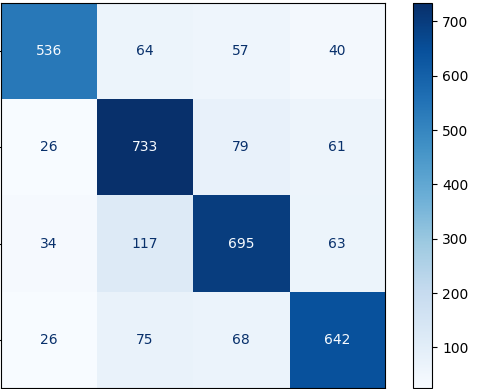}}
\subfloat[LSTM-GRU \label{multi_class_emotion_cmb}]{
\includegraphics[width=0.49\columnwidth]{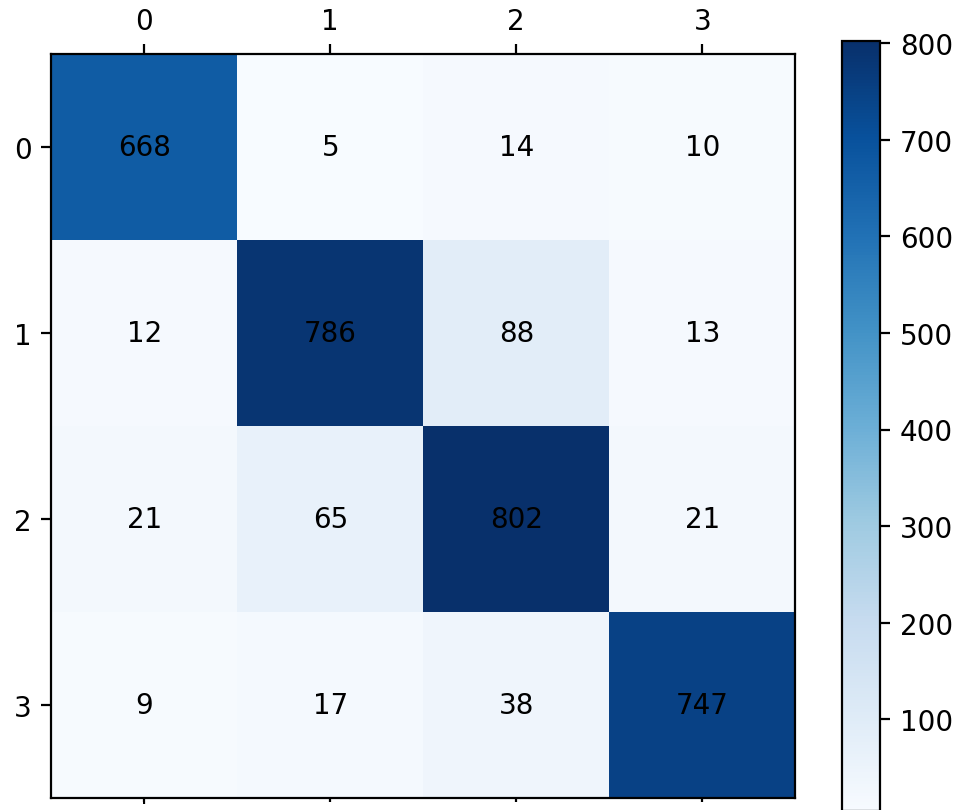}}
\caption{Confusion matrices for Multi-class Emotion Classification}
\label{fig:multi_class_emotion_cm}
\end{figure}

For the multi-class task, the confusion matrix of Random Forest (Fig.~\ref{multi_class_emotion_cma}) shows moderate classification accuracy, particularly excelling in certain classes, such as "funny" and "calm," but struggles with overlapping emotion scores due to the inherent ambiguity and class imbalance in higher label bins. In contrast, the LSTM-GRU model (Fig.~\ref{multi_class_emotion_cmb}) consistently preserves high fidelity in emotion intensity discrimination, showing dominant diagonal entries with minimal spillover, thus validating its superior generalization capacity in complex affective modeling.

\begin{figure}[!ht]
\centering
\subfloat[Random Forest\label{multi_label_cma}]{
\includegraphics[width=0.49\columnwidth]{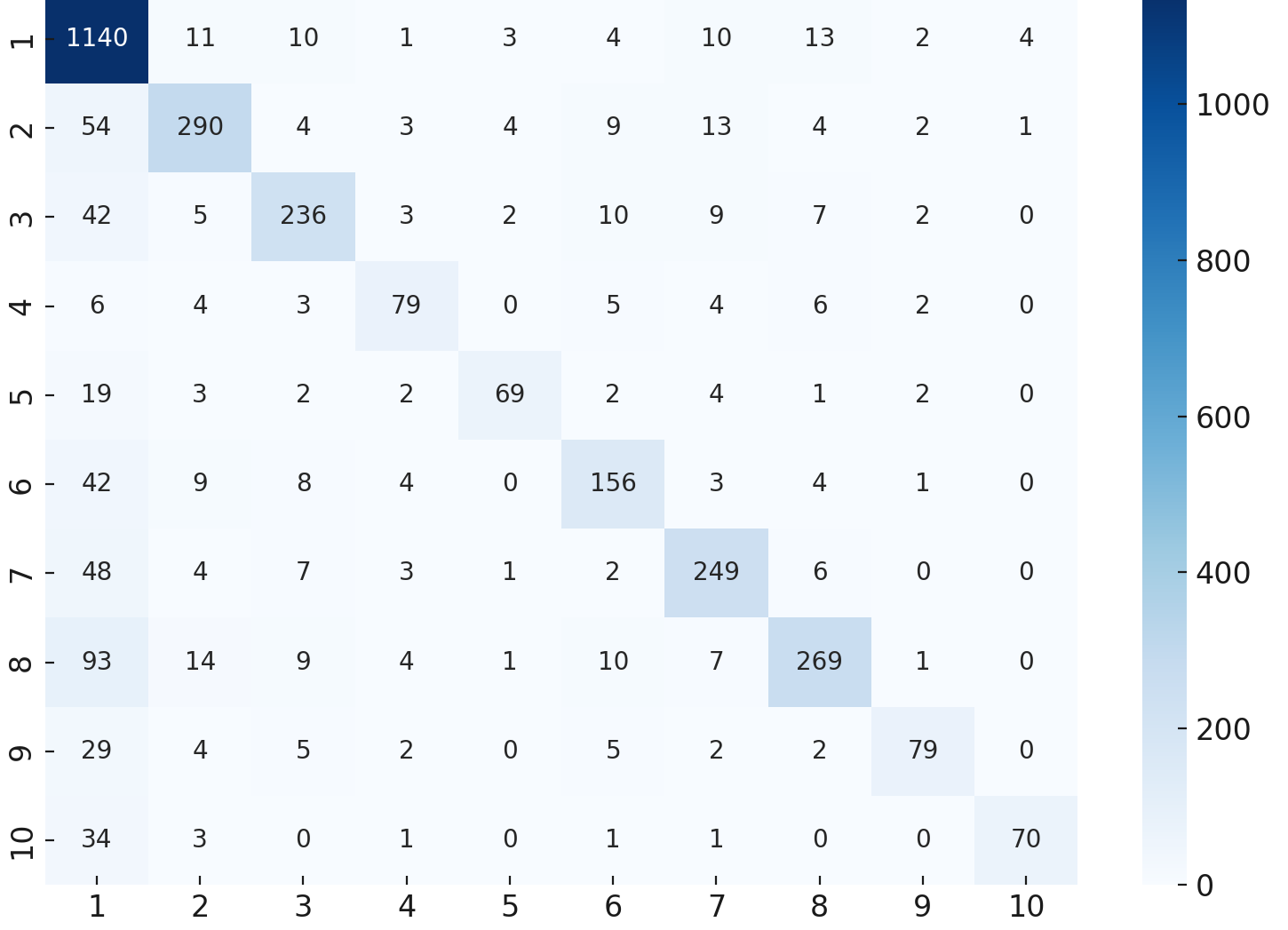}}
\subfloat[LSTM-GRU\label{multi_label_cmb}]{
\includegraphics[width=0.49\columnwidth]{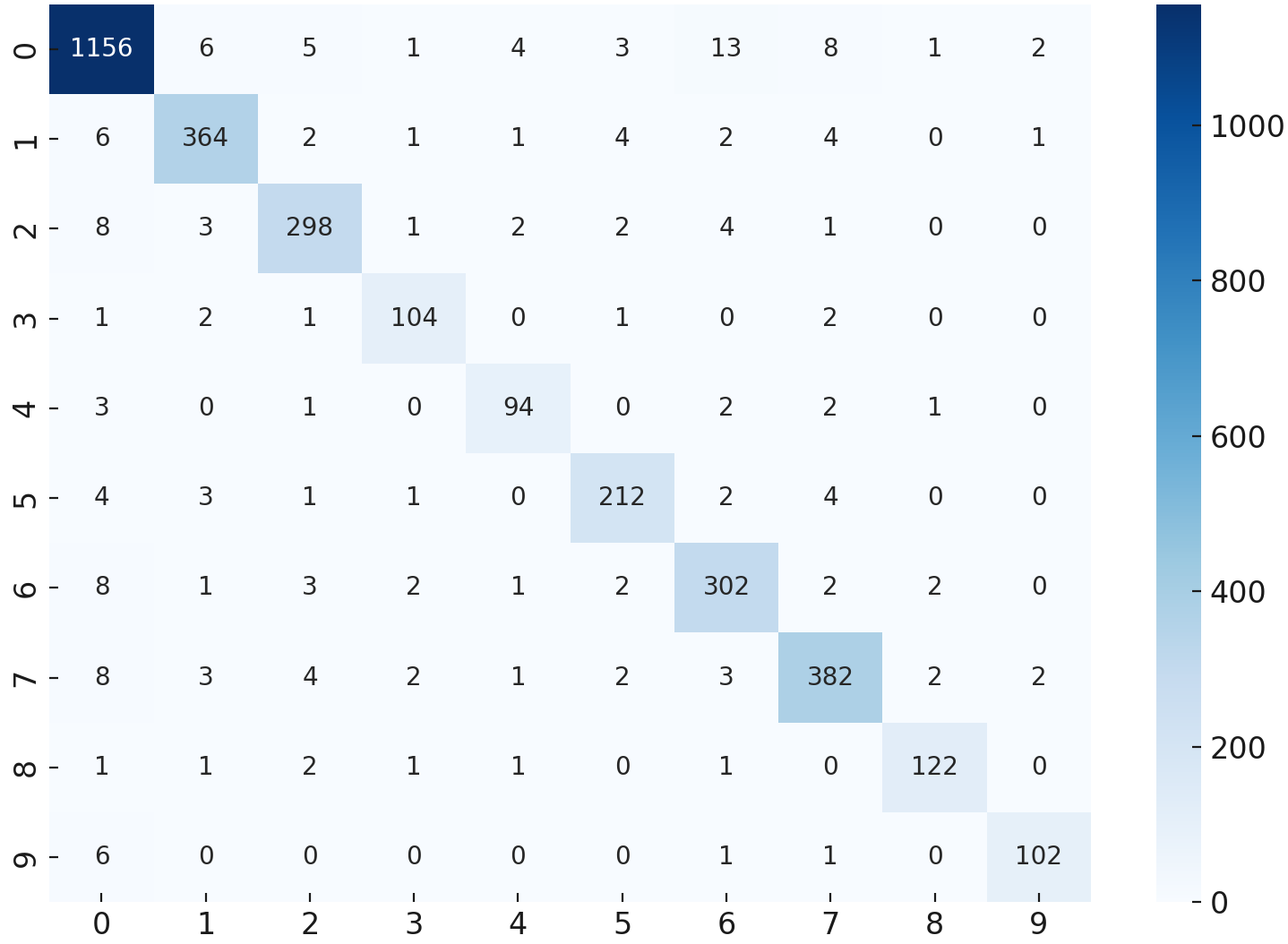}}
\caption{Confusion matrices for Multi-label Classification.}
\label{fig:multi_label_cm}
\end{figure}

Additionally, the Multi-Label Emotion classification results further illustrate the disparity between classical and deep learning paradigms. The confusion matrix for Random Forest (Fig.~\ref{multi_label_cma}) shows that while it performs reasonably across classes such as "funny" and "calm," it exhibits increased confusion between negative affective states like "boring" and "horrible," suggesting limited discriminability. In contrast, the LSTM-GRU model (Fig.~\ref{multi_label_cmb}) demonstrates a clearly defined and balanced diagonal structure with substantially fewer cross-class mispredictions, indicating a strong capacity for jointly learning all four affective labels with temporal consistency and high inter-class separability. Overall, these comparative insights reinforce the effectiveness of temporal deep learning models in EEG-based emotion recognition while acknowledging the interpretability and simplicity of ensemble-based classical approaches.

\begin{figure}[!ht]
\centering
\subfloat[Binary Classification Accuracy Curve \label{fig.3(a)}]{
\includegraphics[width=0.49\columnwidth]{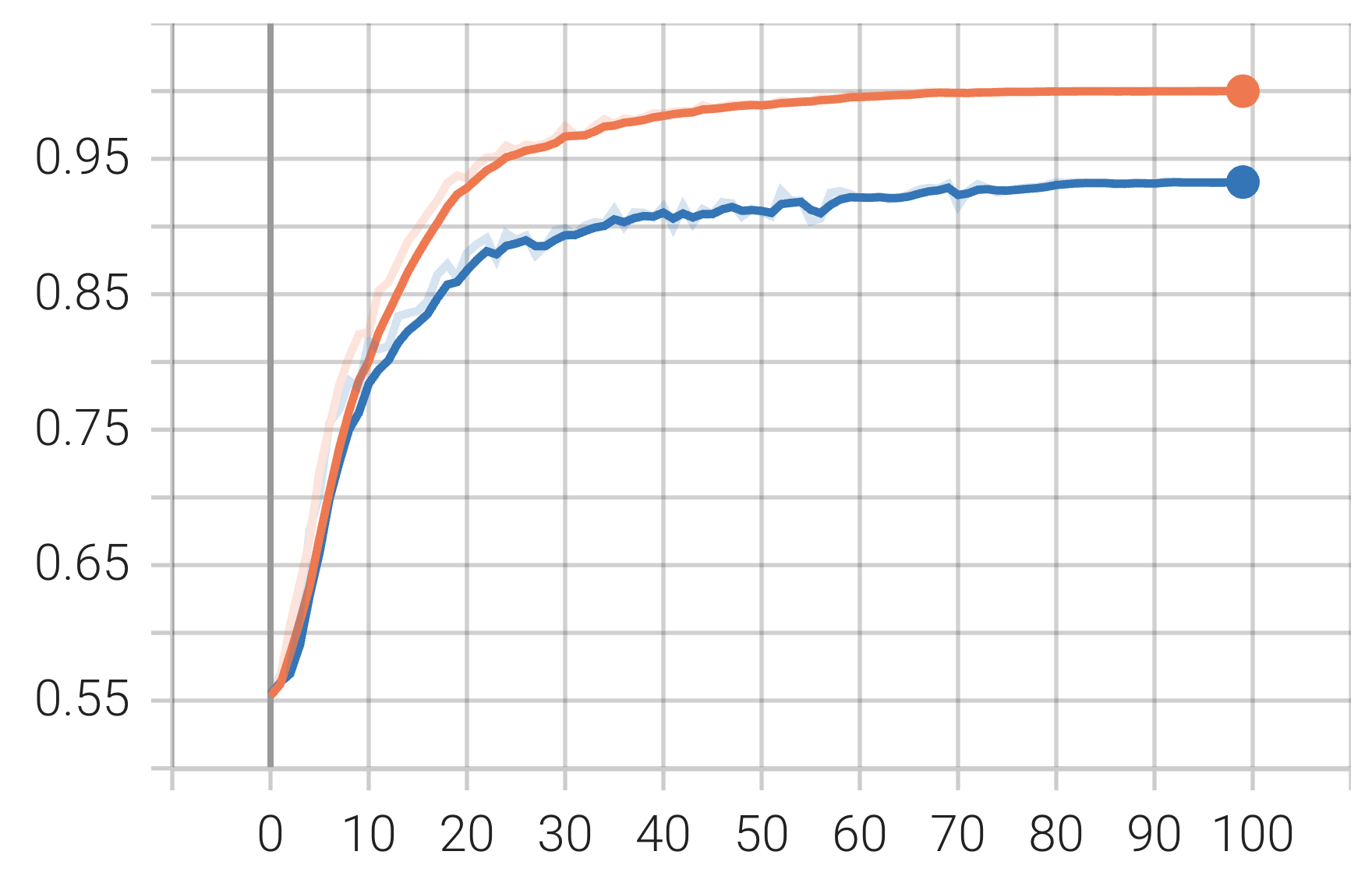}}
\subfloat[Binary Classification Loss Curve\label{fig.3(b)}]{
\includegraphics[width=0.49\columnwidth]{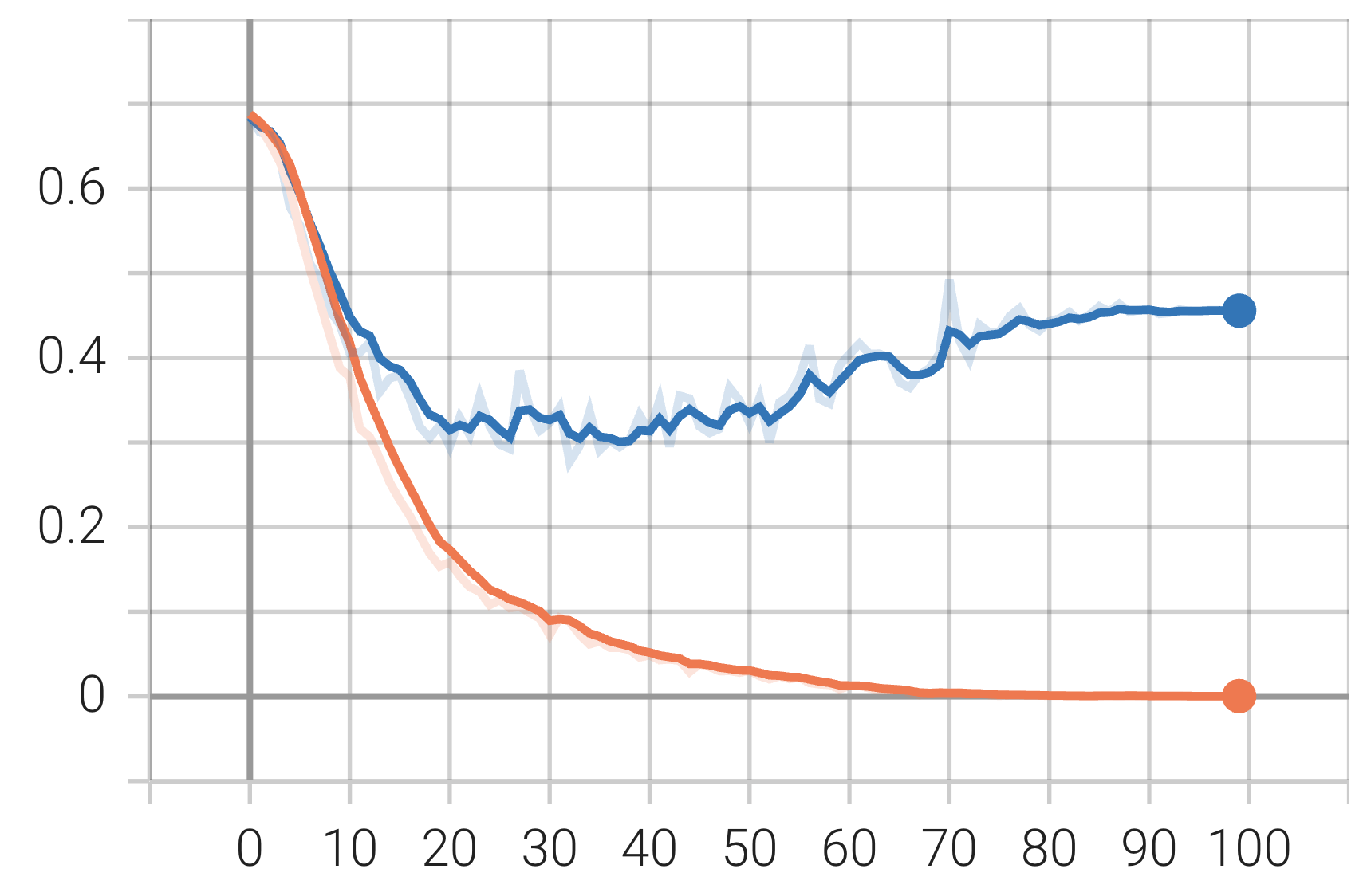}}
\caption{Binary Classification}
\label{fig:binary_classification}
\end{figure}

This paragraph provides the training dynamics of the LSTM-GRU model across binary, multi-class, and Multi-Label Emotion EEG-based emotion classification tasks. In Figs.~\ref {fig.3(a)} and ~\ref{fig.3(b)}, the binary valence classification exhibits rapid and stable convergence, with validation accuracy approaching 93\% and loss decreasing sharply before plateauing, indicating effective learning of emotional polarity.

\begin{figure}[!ht]
\centering
\subfloat[Multi-class Classification Accuracy Curve \label{fig.3(c)}]{
\includegraphics[width=0.49\columnwidth]{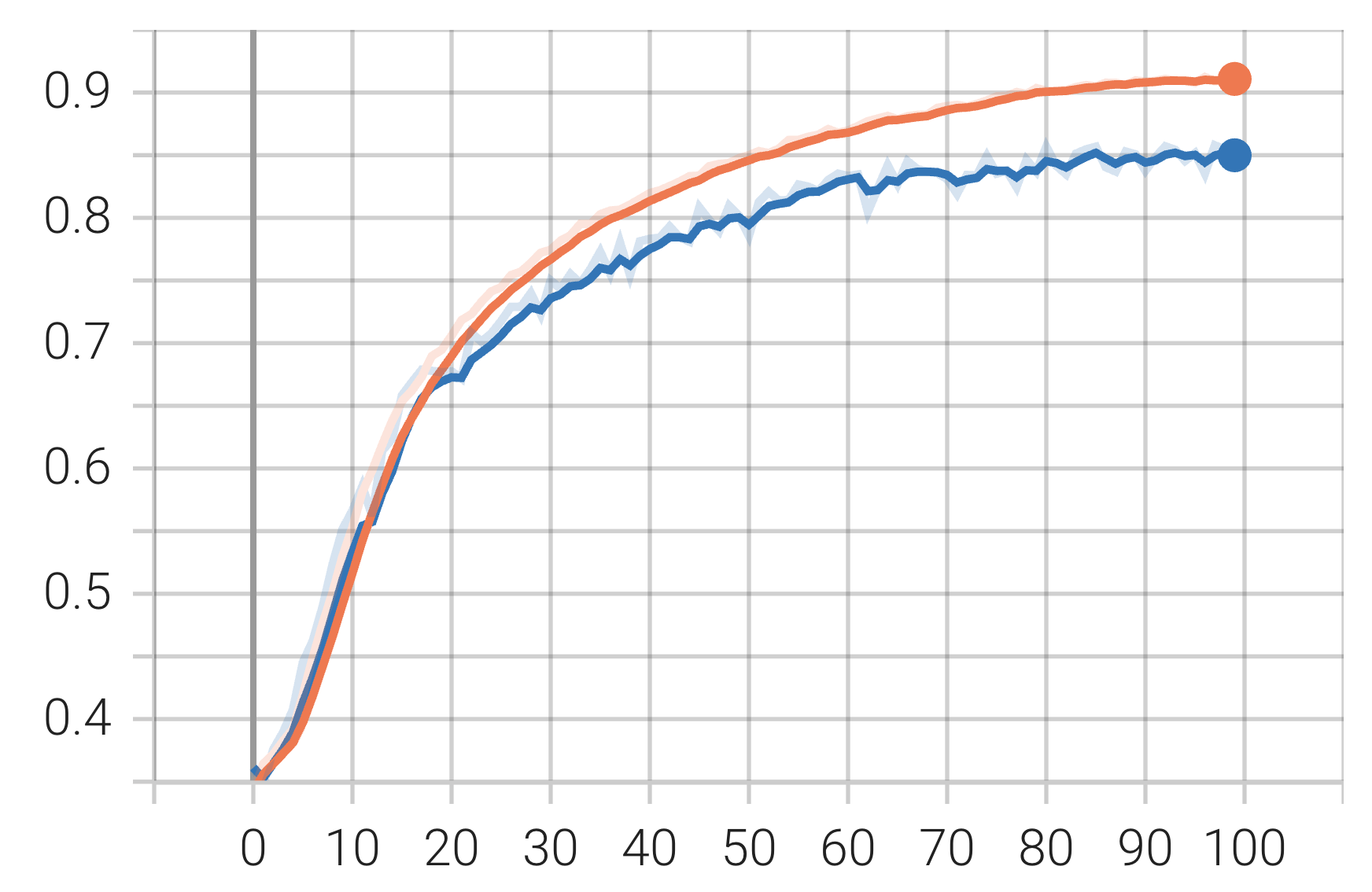}}
\subfloat[Multi-class Classification Loss Curve \label{fig.3(d)}]{
\includegraphics[width=0.49\columnwidth]{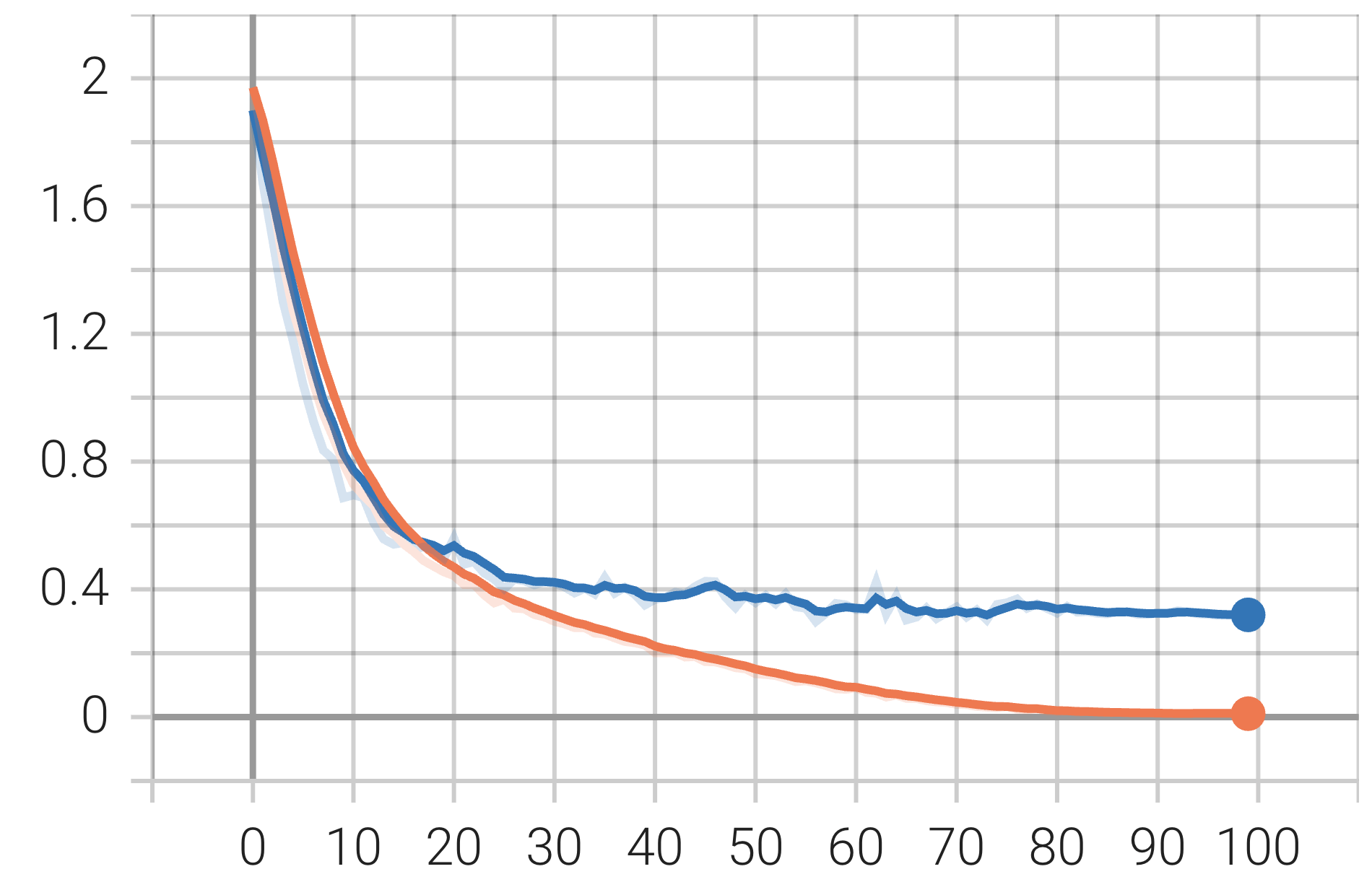}}
\caption{Multi-class Classification}
\label{fig:multi_class_classification}
\end{figure}

Fig.~\ref{fig.3(c)}  and ~\ref{fig.3(d)}  highlight the model’s performance in multi-class classification, where it independently predicts the presence of each emotion (boring, horrible, calm, funny). Here, the accuracy curves steadily improve across all labels, surpassing 92\% in most cases, while the loss curve reflects a smooth decline, suggesting strong generalization despite label co-occurrence.

\begin{figure}[!ht]
\centering
\subfloat[Multi-Label Emotion Classification Accuracy \label{fig.3(e)}]{
\includegraphics[width=0.49\columnwidth]{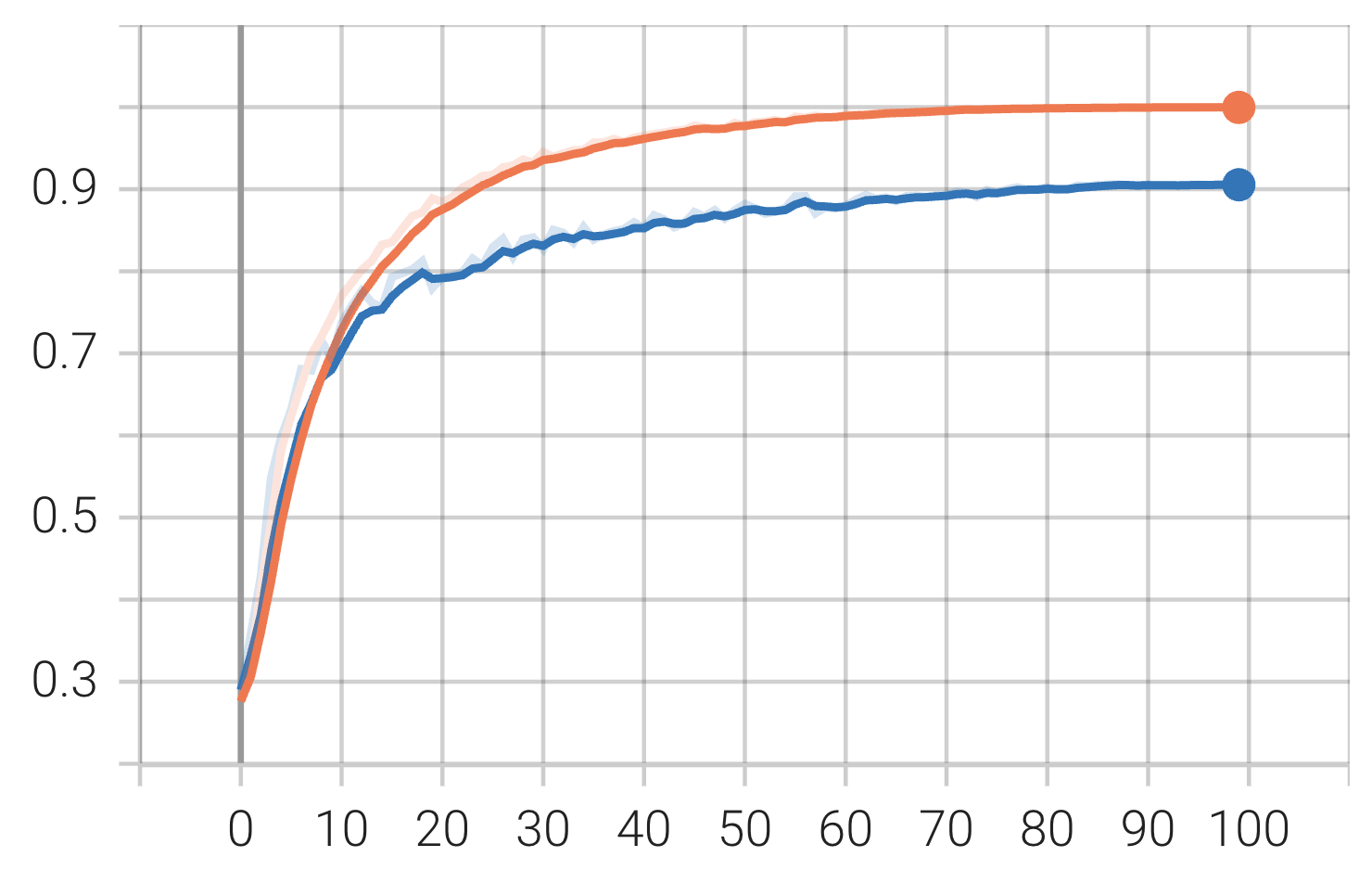}}
\subfloat[Multi-Label Emotion Classification Loss \label{fig.3(f)}]{
\includegraphics[width=0.49\columnwidth]{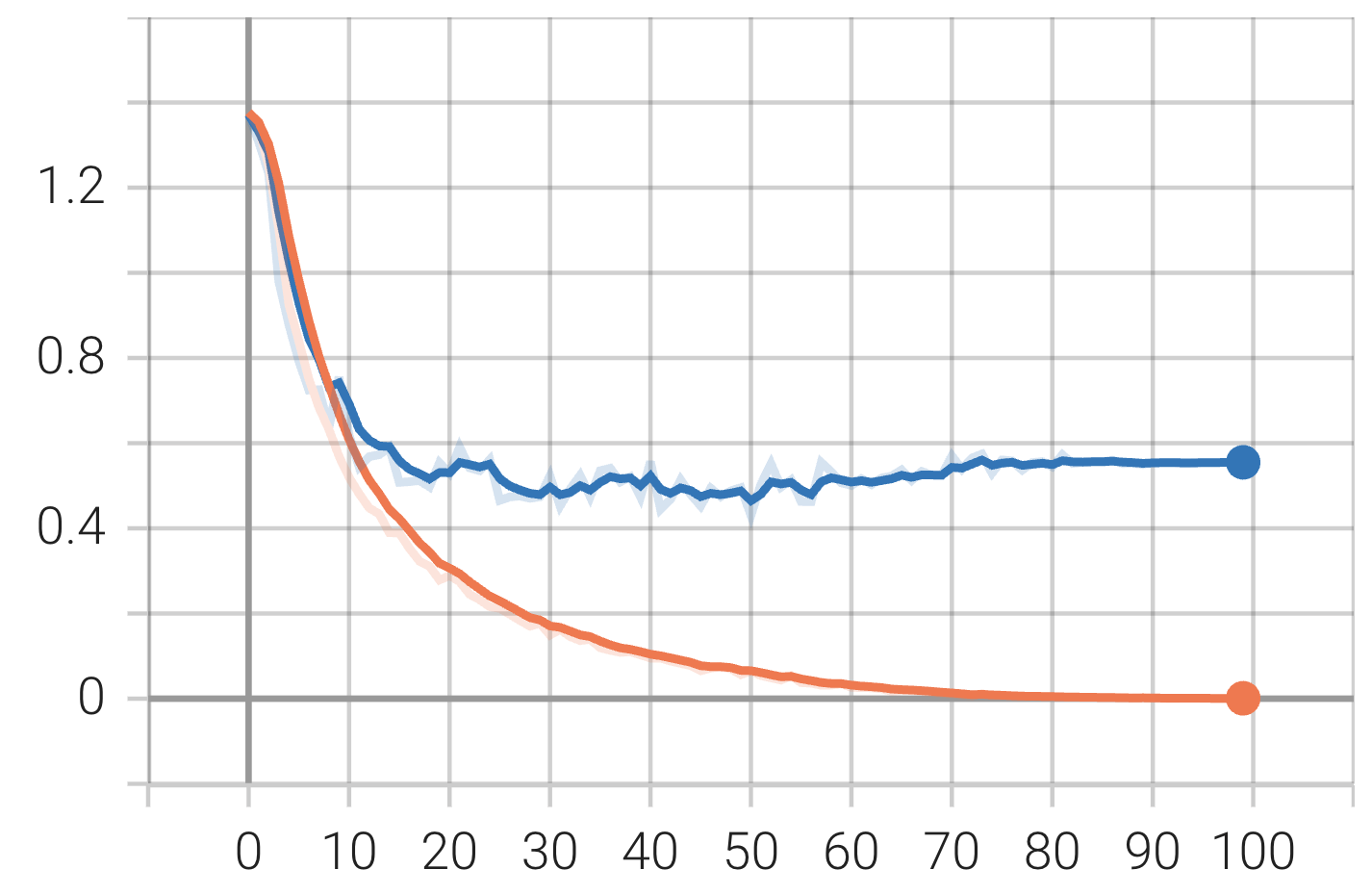}}
\caption{Multi-Label Emotion Classification}
\label{fig:multi_label_emotion_classification}
\end{figure}

In the more complex Multi-Label Emotion classification task depicted in Fig.~\ref{fig.3(e)}  and ~\ref{fig.3(f)}, where a single label is assigned from four exclusive emotion classes per sample, the model continues to demonstrate robust learning behavior, maintaining accuracy beyond 90\% with consistent and stable loss reduction. These results collectively affirm the model’s temporal learning capability and adaptability to varying emotional modeling strategies, from coarse binary distinctions to fine-grained and joint emotion predictions.

\section{Discussion and Comparison} \label{Discussion}
The proposed EEG-based emotion recognition framework demonstrated strong classification performance across binary, multi-class, and Multi-Label Emotion tasks, highlighting the synergy between structured preprocessing and deep temporal modeling. The preprocessing pipeline, comprising systematic parsing of raw EEG recordings, overlapping temporal window segmentation, and extraction of handcrafted statistical and spectral features, significantly enhanced temporal resolution and noise robustness, providing rich input representations for downstream learning. Emotion labels, derived from participants' self-assessed ratings across four affective dimensions (boring, horrible, calm, funny), were encoded into three distinct classification formats: (i) binary valence classification based on the averaged polarity of positive (funny, calm) and negative (boring, horrible) emotion ratings, and (ii) Multi-class emotion classification, where the presence of the most affective state is predicted. (iii) Fine-grained multi-label representation via binning each emotion into 10 ordinal classes, enabling comprehensive modeling of affective states. In binary classification, the LSTM-GRU model achieved a notable 93.3\% accuracy and an F1-score of 0.932, significantly outperforming the Random Forest baseline (85\% accuracy), which validates its capability to capture valence-relevant EEG dynamics. Similarly, in the multi-class task, LSTM-GRU surpassed all classical models, achieving F1-scores exceeding 0.927 across all emotion categories, and offering robust detection of co-occurring emotional states, including high-arousal responses such as "funny." In the more complex Multi-Label Emotion setup, where a single dominant emotion is predicted per instance, the model continued to demonstrate high accuracy (~90.6\%) and reliable generalization, outperforming Random Forest (79\%). Confusion matrix analysis revealed sharper diagonals and reduced misclassifications for LSTM-GRU in all three tasks, especially in distinguishing similar affective states. Training curves further confirmed stable optimization with smooth convergence and no overfitting across tasks. Collectively, these results underscore the effectiveness of combining temporally aware deep models with rigorous preprocessing, supporting a robust and generalizable EEG-based framework for real-time emotion recognition in interactive and neuroadaptive applications.

Table \ref{tab:comparison} presents a comparative analysis of studies, providing an overview of how they analyze emotion recognition using EEG data and related techniques. Shahzad et al. \cite{shahzad2024eeg} and Gosala et al. \cite{gosala2024hybrid} proposed random forest and hybrid CNN for emotion recognition, achieving an accuracy of 98.21\% and 95\%, respectively. On the other hand, the proposed framework achieves 86\% accuracy for binary and 99\% accuracy for multi-class classification problems, while demonstrating the versatility of using transfer learning and outperforming previously conducted works.
\begin{table}[!ht]
\centering
\begin{tabular}{|p{1.7cm}|p{1.4cm}|l|p{1.4cm}|p{1cm}|}
\hline
\textbf{Paper \& Year} & \textbf{Approach} & \textbf{Dataset} & \textbf{Classes} & \textbf{Findings} \\ \hline
Shahzad et al. (2024) \cite{shahzad2024eeg} & Random Forest for Emotion Recognition 
 & GAMEEMO & Multi-class & 98.21\% \\ \hline
Gosala et al. \cite{gosala2024hybrid}& Hybrid CNNs for emotion recognition& GAMEEMO & Multi-class & 95\%\\ \hline
\textbf{Proposed Framework} & Transfer learning & GAMEEMO& Binary, multi-class, multi-label classification using sliding window & 93.3\%, 94.5\%, 90.6\%\\ \hline
\end{tabular}
\caption{Comparison of proposed framework with existing Studies}
\label{tab:comparison}
\end{table}
\section{Conclusion and Future Work}\label{con}
This work introduced a robust and extensible multi-branch EEG-based emotion recognition framework tailored to decode affective responses elicited during game-based cognitive experiences, using the GAMEEMO dataset as a foundational testbed. At its core lies a meticulously engineered preprocessing pipeline that encompasses the synchronized acquisition of 14-channel EEG signals, the extraction of self-reported emotion labels, overlapping temporal windowing, and comprehensive handcrafted feature transformation across both statistical and frequency domains. The framework’s strength stems from its tri-format label encoding strategy: (i) binary valence classification based on the averaged polarity of positive (funny, calm) and negative (boring, horrible) emotion ratings, and (ii) Multi-class emotion classification, where the presence of the most affective state is predicted. (iii) Fine-grained multi-label representation via binning each emotion into 10 ordinal classes, thereby supporting comprehensive benchmarking of learning paradigms. Classical models, such as Random Forest, offered competitive performance with 85\% binary accuracy and macro F1-scores of nearly 0.79 for Multi-Label Emotion classification, demonstrating interpretability and robustness despite class imbalance. However, the LSTM-GRU deep sequence model consistently outperformed all baselines, achieving a 0.932 F1-score in binary tasks and exceeding 0.92 across multi-class and Multi-Label Emotion setups, aided by its temporal modeling of EEG sequences. Confusion matrix analysis confirmed LSTM-GRU’s superior discriminative power, showing dense diagonals and minimal off-class predictions in all classification schemes. Training curves further validated the smooth convergence and strong generalization achieved through effective regularization (dropout, weight decay) and scheduler tuning. Notwithstanding its success, the framework opens up promising future directions, such as integrating domain-aware spectral descriptors (e.g., entropy, wavelets), enhancing subject-independent generalization through meta-learning or transfer learning, and deploying real-time emotion inference on lightweight edge or BCI platforms. These enhancements can pave the way toward scalable, interpretable, and emotionally intelligent neuroadaptive systems in gaming and beyond.
\bibliographystyle{ieeetr}
\bibliography{ref}
\begin{IEEEbiography}[]{Abdul Rehman} is currently pursuing the Ph.D. degree at Western Norway University of Applied Sciences, Bergen, Norway. His research interests include affective computing, EEG-based emotion recognition, deep learning, and the application of serious games in education and healthcare. 
\end{IEEEbiography}

\begin{IEEEbiography}[]{Ilona Heldal} is a professor at Western Norway University of Applied Sciences, aiming to define basic requirements and to develop a better understanding for more effective and enjoyable collaboration via new technologies, especially by utilizing visualization and telecommunication to support work.
\end{IEEEbiography}

\begin{IEEEbiography}[]{Jerry Chun-Wei Lin}
Jerry Chun-Wei Lin (Senior Member, IEEE) received the Ph.D. degree from the Department of Computer Science and Information Engineering, National Cheng Kung University, Tainan, Taiwan, in 2010. He has published over 600 research articles in top-tier, refereed journals and conferences. His research interests encompass data mining, soft computing, artificial intelligence, social computing, multimedia and image processing, as well as privacy-preserving and security technologies.
\end{IEEEbiography}

\end{document}